 \documentclass[final,5p,times,twocolumn]{elsarticle}

\usepackage{amssymb}
\usepackage{amsthm}
\usepackage{times}
\usepackage{epsfig}
\usepackage{graphicx}
\usepackage{amsmath}
\usepackage{amssymb}
\usepackage{url}
\usepackage{comment}
\usepackage{subfigure}
\usepackage{lineno}

\journal{Neurocomputing}

\begin{document}

\begin{frontmatter}



\title{Blur Robust Optical Flow using Motion Channel}


\author[label1,label5]{Wenbin Li}
\author[label2]{Yang Chen}
\author[label3]{JeeHang Lee}
\author[label4]{Gang Ren}
\author[label5]{Darren Cosker}

\address[label1]{Department of Computer Science, University College London, UK}
\address[label2]{Hamlyn Centre, Imperial College London, UK}
\address[label3]{Department of Computer Science, University of Bath, UK}
\address[label4]{School of Digital Art, Xiamen University of Technology, China}
\address[label5]{Centre for the Analysis of Motion, Entertainment Research and Applications (CAMERA), University of Bath, UK}

\begin{abstract}

It is hard to estimate optical flow given a realworld video sequence with camera shake and other motion blur. In this paper, we first investigate the blur parameterization for video footage using near linear motion elements. We then combine a commercial 3D pose sensor with an RGB camera, in order to film video footage of interest together with the camera motion. We illustrates that this additional camera motion/trajectory channel can be embedded into a hybrid framework by interleaving an iterative blind deconvolution and warping based optical flow scheme. Our method yields improved accuracy within three other state-of-the-art baselines given our proposed ground truth blurry sequences; and several other realworld sequences filmed by our imaging system.

\end{abstract}


\end{frontmatter}


\section{Introduction}

Optical flow estimation has been widely applied to computer vision applications, e.g. segmentation, image deblurring and stabilization, etc. In many cases, optical flow is often estimated on the videos captured by a shaking camera. Those footages may contain a significant amount of camera blur that bring additional difficulties into the traditional variational optical flow framework. It is because such blur scenes often lead to a fact that a pixel may match multiple pixels between image pair. It further violates the basic assumption -- intensity constancy -- of the optical flow framework.

\begin{figure}[t!]
\centerline{
\includegraphics[width=0.9\linewidth]{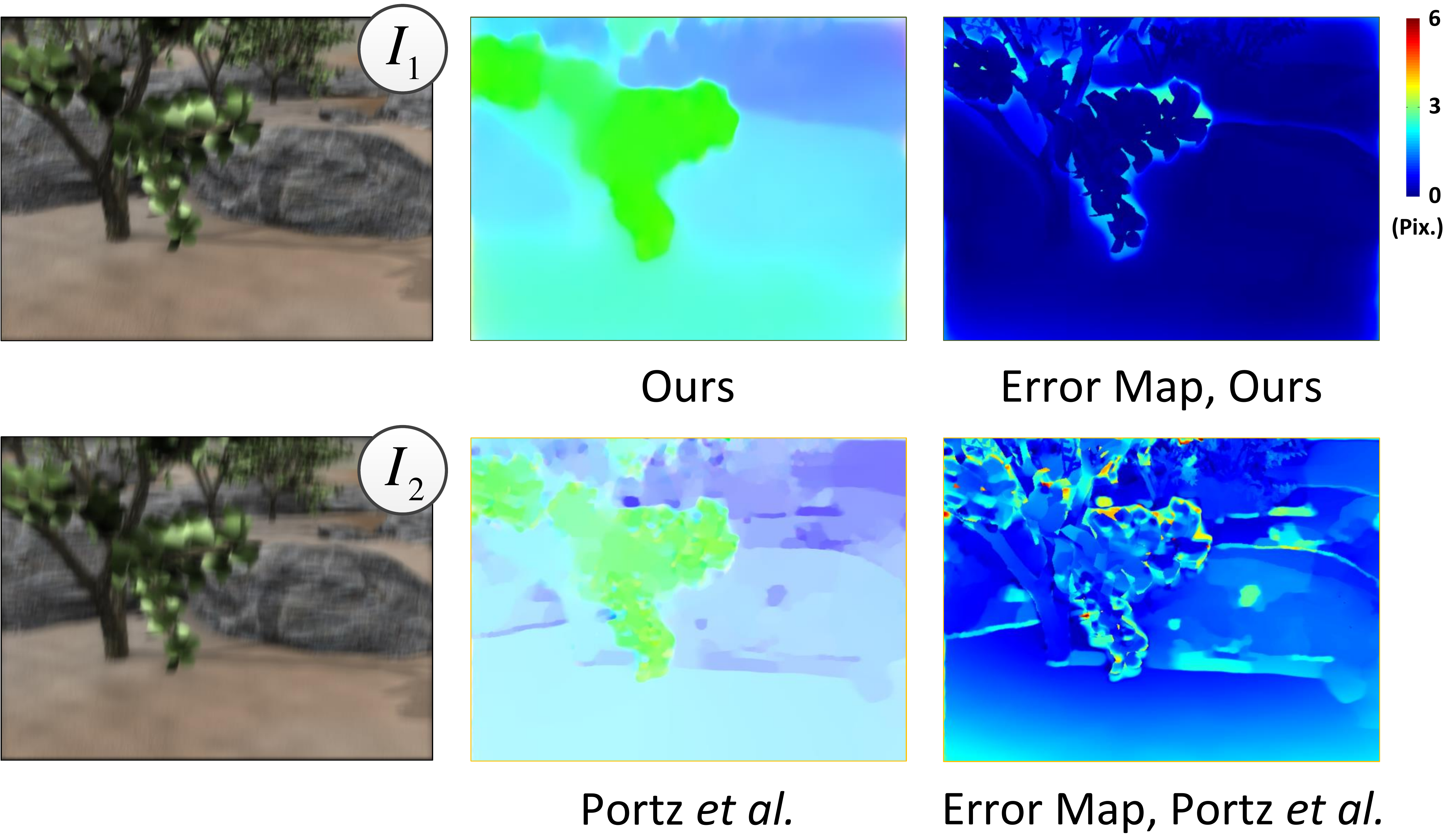}
}
\caption{Visual comparison of our method to Portz~\emph{et al.}~\cite{Portz} on our ground truth benchmark \emph{Grove2} with synthetic camera shake blur. \textbf{First Column}: the input images; \textbf{Second Column}: the optical flow fields calculated by our method and the baseline; \textbf{Third Column}: the RMS error maps against the ground truth.}
\label{moBlur:fig:firstFig}
\end{figure}

In this paper, we investigate the issue of how to precisely estimate optical flow from a blurry video footage. We observe that the blur kernel between neighboring frames may be near linear, which can be parameterized using linear elements of the camera motion. In this case, the camera trajectory can be informatic to enhance the image deblurring within a variational optical flow framework. Based on this observation, our major contribution in this paper is to utilise an \emph{RGB-Motion Imaging System} -- an RGB sensor combined with a 3D pose\&position tracker -- in order to propose: \textbf{(A)} an iterative enhancement process for camera shake blur estimation which encompasses the tracked camera motion (Sec.~\ref{moBlur:sec:system}) and a \emph{Directional High-pass Filter} (Sec.~\ref{moBlur:sec:filter} and Sec.~\ref{moBlur:sec:filtering});  \textbf{(B)} a \emph{Blur-Robust Optical Flow Energy} formulation (Sec.~\ref{moBlur:sec:robustEng}); and \textbf{(C)} a hybrid coarse-to-fine framework (Sec.~\ref{moBlur:sec:oflowFramework}) for computing optical flow in blur scenes by interleaving an iterative blind deconvolution process and a warping based minimisation scheme. In the evaluation section, we compare our method to three existing state-of-the-art optical flow approaches on our proposed ground truth sequences (Fig.~\ref{moBlur:fig:firstFig}, blur and baseline blur-free equivalents) and also illustrate the practical benefit of our algorithm given realworld cases.

\vspace{-1mm}
\section{Related Work}
\vspace{-1mm}

Camera shake blur often occurs during fast camera movement in low-light conditions due to the requirement of adopting a longer exposure. Recovering both the blur kernel and the latent image from a single blurred image is known as \emph{Blind Deconvolution} which is an inherently ill-posed problem. Cho and Lee~\cite{FMD} propose a fast deblurring process within a coarse-to-fine framework (Cho\&Lee) using a predicted edge map as a prior. To reduce the noise effect in this framework, Zhong \emph{et al.}~\cite{Zhong} introduce a pre-filtering process which reduces the noise along a specific direction and preserves the image information in other directions. Their improved framework provides high quality kernel estimation with a low run-time but shows difficulties given combined object and camera shake blur.

\begin{figure*}[t!]
\centerline{
\includegraphics[width=1.02\linewidth]{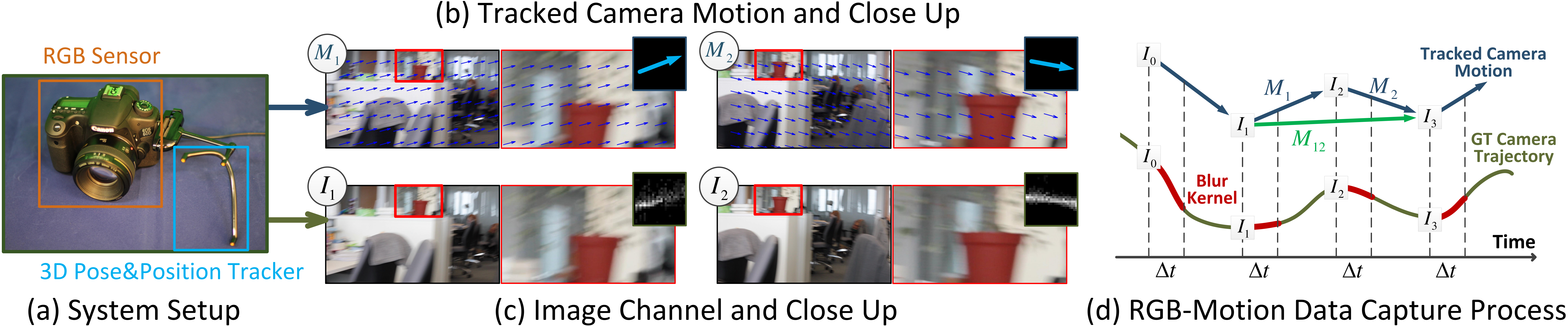}
}
\caption{RGB-Motion Imaging System. \textbf{(a)}: Our system setup using a combined RGB sensor and 3D Pose\&Position Tracker. \textbf{(b)}: The tracked 3D camera motion in relative frames. The top-right box is the average motion vector -- which has similar direction to the blur kernel. \textbf{(c)}:  Images captured from our system. The top-right box presents the blur kernel estimated using ~\cite{FMD}. \textbf{(d)}: The internal process of our system where the $\Delta t$ presents the exposure time.}
\label{moBlur:fig:camTrajectory}
\end{figure*}

To obtain higher performance, a handful of combined hardware and software-based approaches have also been proposed for image deblurring. Tai \emph{et al.}~\cite{tai08} introduce a hybrid imaging system that is able to capture both video at high frame rate and a blurry image. The optical flow fields between the video frames are utilised to guide blur kernel estimation. Levin \emph{et al.}~\cite{levin} propose to capture a uniformly blurred image by controlling the camera motion along a parabolic arc. Such uniform blur can then be removed based on the speed or direction of the known arc motion. As a complement to Levin \emph{el al.}'s~\cite{levin} hardware-based deblurring algorithm, Joshi \emph{et al.}~\cite{Joshi} apply inertial sensors to capture the acceleration and angular velocity of a camera over the course of a single exposure. This extra information is introduced as a constraint in their energy optimisation scheme for recovering the blur kernel. All the hardware-assisted solutions described provide extra information in addition to the blurry image, which significantly improves overall performance. However, the methods require complex electronic setups and the precise calibration.

Optical flow techniques are widely studied and adopted across computer vision because of dense image correspondences they provide. Such dense tracking is important for other fundamental research topics e.g. 3D reconstruction~\cite{reflection} and visual effects~\cite{lv2013game,lv2014multimodal}, etc. In the last two decades, the optical flow model has evolved extensively -- one landmark work being the variational model of Horn and Schunck~\cite{HS} where the concept of \emph{Brightness Constancy} is proposed. Under this assumption, pixel intensity does not change spatio-temporally, which is, however, often weakened in realworld images because of natural noise. To address this issue, some complementary concepts have been developed to improve performance given large displacements~\cite{Brox}, taking advantage of feature-rich surfaces~\cite{Xu_deblur} and adapting to nonrigid deformation in scenes~\cite{APO,LME,APO_JIFS,moBlur,tang,li2013nonrigid}. However, flow approaches that can perform well given blurred scenes -- where the \emph{Brightness Constancy} is usually violated -- are less common. Of the approaches that do exist, Schoueri \emph{et al.}~\cite{Schoueri} perform a linear deblurring filter before optical flow estimation while Portz \emph{et al.}~\cite{Portz} attempt to match un-uniform camera motion between neighbouring input images. Whereas the former approach may be limited given nonlinear blur in realworld scenes; the latter requires two extra frames to parameterise the motion-induced blur. Regarding non optical-flow based methods, Yuan \emph{et al.}~\cite{Yuan} align a blurred image to a sharp one by predefining an affine image transform with a blur kernel. Similarly HaCohen \emph{et al.}~\cite{HaCohen} achieve alignment between a blurred image and a sharp one by embedding deblurring into the correspondence estimation. Li \emph{et al.}~\cite{moBlur} present an approach to solve the image deblurring and optical flow simultaneously by using the RGB-Motion imaging.

\section{RGB-Motion Imaging System}
\label{moBlur:sec:system}

Camera shake blur within video footage is typically due to fast camera motion and/or long exposure time. In particular, such blur can be considered as a function of the camera trajectory supplied to image space during the exposure time $\Delta t$. It therefore follows that knowledge of the actual camera motion between image pairs can provide significant information when performing image deblurring~\cite{Joshi,levin}.

In this paper, we propose a simple and portable setup (Fig.~\ref{moBlur:fig:camTrajectory}(a)), combining an RGB sensor and a 3D pose\&position tracker (SmartNav by NaturalPoint Inc.) in order to capture continuous scenes (video footage) along with real-time camera pose\&position information. Note that the RGB sensor could be any camera or a Kinect sensor -- A Canon EOS 60D is applied in our implementation to capture $1920\times1080$ video at frame rate of 24 FPS. Furthermore, our tracker is proposed to provide the rotation (yaw, pitch and roll), translation and zoom information within a reasonable error range (2 mm). To synchronise this tracker data and the image recording, a real time collaboration (RTC) server~\cite{JeeHang} is built using the instant messaging protocol XMPP (also known as Jabber\footnote{http://www.jabber.org/}) which is designed for message-oriented communication based on XML, and allows real-time responses between different messaging channels or any signal channels that can be transmitted and received in message form. In this case, a time stamp is assigned to the received message package by the central timer of the server. Those message packages are synchronised if they contain nearly the same time stamp. We consider the Jabber for synchronisation because of its opensource nature and the low respond delay (around 10 ms).

Assuming objects have similar depth within the same scene (a common assumption in image deblurring which will be discussed in our future work), the tracked 3D camera motion in image coordinates can be formulated as:

\begin{align}
\textbf{M}_j=\frac{1}{n}\sum_{\textbf{x}}K \left ( \left [ R|T \right ]\textbf{X}_{j+1}-\textbf{X}_{j}\right )
\end{align}

where $\textbf{M}_j$ represents the average of the camera motion vectors from the image $j$ to image $j+1$. $\textbf{X}$ denotes the 3D position of the camera while $\textbf{x}=(x,y)^T$ is a pixel location and $n$ represents the number of pixels in an image. $K$ represents the 3D projection matrix while $R$ and $T$ denote the rotation and translation matrices respectively of tracked camera motion in the image domain. All these information $K$, $R$ and $T$ is computed using \texttt{Optitrack}'s \emph{Camera SDK}\footnote{http://www.naturalpoint.com/optitrack} (version 1.2.1). Fig~\ref{moBlur:fig:camTrajectory}(b,c) shows sample data (video frames and camera motion) captured from our imaging system. It is observed that blur from the realworld video is near linear due to the relatively high sampling rate of the camera. The blur direction can therefore be approximately described using the tracked camera motion. Let the tracked camera motion $\textbf{M}_j=(r_j,\theta_j)^T$ be represented in polar coordinates where $r_j$ and $\theta_j$ denote the magnitude and directional component respectively. $j$ is a sharing index between tracked camera motion and frame number. In addition, we also consider the combined camera motion vector of neighbouring images as shown in Fig~\ref{moBlur:fig:camTrajectory}(d), e.g. $\textbf{M}_{12}=\textbf{M}_1+\textbf{M}_2$ where $\textbf{M}_{12}=(r_{12},\theta_{12})$ denotes the combined camera motion vector from image 1 to image 3. As one of our main contributions, these real-time motion vectors are proposed to provide additional constraints for blur kernel enhancement (Sec.~\ref{moBlur:sec:oflowFramework}) within our framework.

\section{Blind Deconvolution}
\label{moBlur:sec:filter}

The motion blur process can commonly be formulated:

{\setlength\abovedisplayskip{-1mm}
\setlength\belowdisplayskip{-1mm}
\begin{align}
I = k \otimes l +n
\end{align}
}

where $I$ is a blurred image and $k$ represents a blur kernel w.r.t. a specific \emph{Point Spread Function}. $l$ is the latent image of $I$; $\otimes$ denotes the convolution operation and $n$ represents spatial noise within the scene. In the blind deconvolution operation, both $k$ and $l$ are estimated from $I$, which is an ill-posed (but extensively studied) problem. A common approach for blind deconvolution is to solve both $k$ and $l$ in an iterative framework using a coarse-to-fine strategy:

{\setlength\abovedisplayskip{-1mm}
\setlength\belowdisplayskip{-1mm}
\begin{align}
k &= \textbf{argmin}_k\{ \left \| I-k\otimes l \right \|+\rho(k)\},\\
l &= \textbf{argmin}_l\{ \left \| I-k\otimes l \right \|+\rho(l)\}.
\end{align}
}

where $\rho$ represents a regularization that penalizes spatial smoothness with a sparsity prior~\cite{FMD}, and is widely used in recent state-of-the-art work~\cite{Shan,Xu_deblur}. Due to noise sensitivity, low-pass and bilateral filters~\cite{Tai} are typically employed before deconvolution. Eq.~\ref{moBlur:eq:blurKernel} denotes the common definition of an optimal kernel from a filtered image.

{\setlength\abovedisplayskip{-1mm}
\setlength\belowdisplayskip{-1mm}
\begin{align}
k_f &= \textbf{argmin}_{k_f}\{ \left \| (k \otimes l +n)\otimes f - k_f \otimes l\right \|+\rho(k_f)\} \nonumber\\
&\approx \textbf{argmin}_{k_f} \left \| l\otimes(k \otimes f - k_f) \right \| = k\otimes f
\label{moBlur:eq:blurKernel}
\end{align}
}

where $k$ represents the ground truth blur kernel, $f$ is a filter, and $k_f$ denotes the optimal blur kernel from the filtered image $I\otimes f$. The low-pass filtering process improves deconvolution computation by removing spatially-varying high frequency noise but also results in the removal of useful information which yields additional errors over object boundaries. To preserve this useful information, we introduce a directional high-pass filter that utilises our tracked 3D camera motion.


\section{Directional High-pass Filter}

\begin{figure*}[t!]
\centerline{
\includegraphics[width=0.75\linewidth]{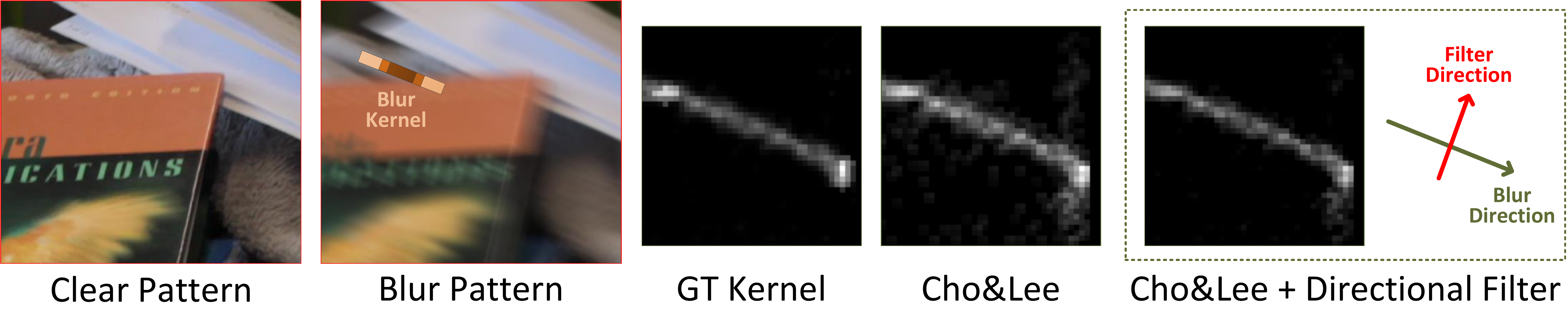}
}
\caption{Directional high-pass filter for blur kernel enhancement. Given the blur direction $\theta$, a directional high-pass filter along $\theta+\pi/2$ is applied to preserve blur detail in the estimated blur kernel.}
\label{moBlur:fig:DFsample}
\end{figure*}

Detail enhancement using directional filters has been proved effective in several areas of computer vision~\cite{Zhong}. Here we define a directional high-pass filter as:

{\setlength\abovedisplayskip{-1mm}
\setlength\belowdisplayskip{-1mm}
\begin{align}
f_\theta \otimes I(\textbf{x}) = m\int g(t)I(\textbf{x}+t\Theta)dt
\label{moBlur:eq:filtering}
\end{align}
}

where $\textbf{x}=(x,y)^T$ represents a pixel position and $g(t)=1-exp\{-t^2/2\sigma^2\}$ denotes a 1D Gaussian based high-pass function. $\Theta=(\cos\theta,\sin\theta)^T$ controls the filtering direction along $\theta$. $m$ is a normalization factor defined as $m=\left ( \int g(t)dt\right )^{-1}$. The filter $f_\theta$ is proposed to preserve overall high frequency details along direction $\theta$ without affecting blur detail in orthogonal directions~\cite{Chen}. Given a directionally filtered image $b_\theta = f_\theta \otimes I(\textbf{x})$, the optimal blur kernel is defined (Eq~\ref{moBlur:eq:blurKernel}) as $k_\theta = k \otimes f_\theta$. Fig.~\ref{moBlur:fig:DFsample} demonstrates that noise or object motion within a scene usually results in low frequency noise in the estimated blur kernel (Cho\&Lee~\cite{FMD}). This low frequency noise can be removed by our directional high-pass filter while preserving major blur details. In our method, this directional high-pass filter is supplemented into the Cho\&Lee~\cite{FMD} framework using a coarse-to-fine strategy in order to recover high quality blur kernels for use in our optical flow estimation (Sec.~\ref{moBlur:sec:filtering}).

\section{Blur-Robust Optical Flow Energy}
\label{moBlur:sec:robustEng}

Within a blurry scene, a pair of adjacent natural images may contain different blur kernels, further violating \emph{Brightness Constancy}. This results in unpredictable flow error across the different blur regions. To address this issue, Portz \emph{et al.} proposed a modified \emph{Brightness Constancy} term by matching the un-uniform blur between the input images. As one of our main contributions, we extend this assumption to a novel \emph{Blur Gradient Constancy} term in order to provide extra robustness against illumination change and outliers. Our main energy function is given as follows:

{\setlength\abovedisplayskip{-1mm}
\setlength\belowdisplayskip{-1mm}
\begin{align}
\centering
E(\textbf{w}) = E_{B}(\textbf{w}) + \gamma E_{S}(\textbf{w})
\label{moBlur:eq:Eng}
\end{align}
}

A pair of consecutively observed frames from an image sequence is considered in our algorithm. $I_{1}(\textbf{x})$ represents the current frame and its successor is denoted by $I_{2}(\textbf{x})$ where $I_* = k_*\otimes l_*$ and $\{I_{*}, l_*:\Omega \subset \mathbb{R}^3 \to \mathbb{R}\}$ represent rectangular images in the RGB channel. Here $l_*$ is latent image and $k_*$ denotes the relative blur kernel. The optical flow displacement between $I_{1}(\textbf{x})$ and $I_{2}(\textbf{x})$ is defined as $\textbf{w} = (u,v)^{T}$. To match the un-uniform blur between input images, the blur kernel from each input image is applied to the other. We have new blur images $b_1$ and $b_2$ as follows:

{\setlength\abovedisplayskip{-1mm}
\setlength\belowdisplayskip{-1mm}
\begin{align}
b_1 = k_2\otimes I_1 \approx k_2 \otimes k_1 \otimes l_1\\
b_2 = k_1\otimes I_2 \approx k_1 \otimes k_2 \otimes l_2
\label{moBlur:eq:uniformImg}
\end{align}
}

Our energy term encompassing \emph{Brightness} and \emph{Gradient Constancy} relates to $b_1$ and $b_2$ as follows:

{\setlength\abovedisplayskip{-1mm}
\setlength\belowdisplayskip{-1mm}
\begin{align}
E_{B}(\textbf{w}) &= \int_{\Omega} \phi ( \left \| b_{2}(\textbf{x}+\textbf{w}) - b_{1}(\textbf{x})\right \|^{2}\nonumber\\
&+ \alpha \left \| \nabla b_{2}(\textbf{x}+\textbf{w}) - \nabla b_{1}(\textbf{x})\right \|^{2})d\textbf{x}
\end{align}
}

The term $\nabla = (\partial_{xx},\partial_{yy})^{T}$ presents a spatial gradient and $\alpha\in[0,1]$ denotes a linear weight. The smoothness regulariser penalizes global variation as follows:

{\setlength\abovedisplayskip{-1mm}
\setlength\belowdisplayskip{-1mm}
\begin{eqnarray}
E_{S}(\textbf{w}) = \int_{\Omega} \phi(\left \| \nabla u \right \|^{2} + \left \| \nabla v \right \|^{2})d\textbf{x}
\end{eqnarray}
}

\begin{table}[t!]
\centerline{
\begin{tabular}{l}
\hline
\hspace{5 pt}\textbf{Algorithm 1}: Blur-Robust Optical Flow Framework\\
\hline
\hspace{5 pt}\textbf{Input} \hspace{6 pt} :~A image pair $I_1$, $I_2$ and camera motion $\theta_1$, $\theta_2$, $\theta_{12}$\\
\hspace{5 pt}\textbf{Output}~:~Optimal optical flow field $\textbf{w}$\vspace{1pt}\\
\hspace{5 pt}1:\hspace{5 pt} \emph{A $n$-level top-down pyramid is built with the level index $i$}\\
\hspace{5 pt}2:\hspace{5 pt} $i \leftarrow 0$\\
\hspace{5 pt}3:\hspace{5 pt} $l^i_1 \leftarrow I^i_1$, $l^i_2 \leftarrow I^i_2$\\
\hspace{5 pt}4:\hspace{5 pt} $k^i_1 \leftarrow 0$, $k^i_2 \leftarrow 0$, $\textbf{w}^i\leftarrow (0,0)^T$\vspace{1pt}\\
\hspace{5 pt}5:\hspace{5 pt} \textbf{for} \emph{coarse to fine} \textbf{do}\\
\hspace{5 pt}6:\hspace{30 pt} $i \leftarrow i+1$\\
\hspace{5 pt}7:\hspace{30 pt} Resize $k^i_{\{1,2\}}$, $l^i_{\{1,2\}}$, $I^i_{\{1,2\}}$ and $\textbf{w}^i$ with the $i$th scale\\
\hspace{5 pt}8:\hspace{30 pt} \textbf{foreach} $* \in \{1,2\}$ \textbf{do}\\
\hspace{5 pt}9:\hspace{55 pt} $k^i_* \leftarrow$ \texttt{IterBlindDeconv} ( $l^i_*, I^i_*$ )\\
10:\hspace{55 pt} $k^i_* \leftarrow$ \texttt{DirectFilter} ( $k^i_*, \theta_1, \theta_2, \theta_{12}$ )\\
11:\hspace{55 pt} $l^i_* \leftarrow$ \texttt{NonBlindDeconvolve} ( $k^i_*, I^i_*$ )\\
12:\hspace{30 pt} \textbf{endfor}\\
13:\hspace{30 pt} $b^i_1 \leftarrow I^i_1 \otimes k^i_2$, $b^i_2 \leftarrow I^i_2 \otimes k^i_1$\\
14:\hspace{30 pt} $d\textbf{w}^i\leftarrow$ \texttt{Energyoptimisation} ( $b^i_1, b^i_2, \textbf{w}^i$ )\\
15:\hspace{30 pt} $\textbf{w}^i\leftarrow \textbf{w}^i+d\textbf{w}^i$\\
16:\hspace{5 pt} \textbf{endfor}\\
\hline
\label{moBlur:tab:kernelOflow}
\end{tabular}
\vspace{-3mm}
}
\end{table}

where we apply the Lorentzian regularisation $\phi(s)=log(1+s^2/2\epsilon^{2})$ to both the data term and smoothness term. In our case, the image properties, e.g. small details and edges, are broken by the camera blur, which leads to additional errors in those regions. We suppose to apply strong boundary preservation even the non-convex Lorentzian regularisation may bring the extra difficulty to the energy optimisation (More analysis can be found in Li~\emph{et al.}~\cite{LME}). In the following section, our optical flow framework is introduced in detail.

\section{Optical Flow Framework}
\label{moBlur:sec:oflowFramework}

Our overall framework is outlined in Algorithm~1 based on an iterative top-down, coarse-to-fine strategy. Prior to minimizing the \emph{Blur-Robust Optical Flow Energy} (Sec.~\ref{moBlur:sec:optimisation}), a fast blind deconvolution approach~\cite{FMD} is performed for pre-estimation of the blur kernel (Sec.~\ref{moBlur:sec:blindDec}), which is followed by kernel refinement using our \emph{Directional High-pass Filter} (Sec.~\ref{moBlur:sec:filtering}). All these steps are detailed in the following subsections.

\subsection{Iterative Blind Deconvolution}
\label{moBlur:sec:blindDec}

Cho and Lee~\cite{FMD} describe a fast and accurate approach (Cho\&Lee) to recover the unique blur kernel. As shown in Algorithm~1, we perform a similar approach for the pre-estimation of the blur kernel $k$ within our iterative process, which involves two steps of prediction and kernel estimation. Given the latent image $l$ estimated from the consecutively coarser level, the gradient maps $\Delta l = \{\partial_x l,\partial_y l\}$ of $l$ are calculated along the horizontal and vertical directions respectively in order to enhance salient edges and reduce noise in featureless regions of $l$. Next, the predicted gradient maps $\Delta l$ as well as the gradient map of the blurry image $I$ are utilised to compute the pre-estimated blur kernel by minimizing the energy function as follows:

{\setlength\abovedisplayskip{-1mm}
\setlength\belowdisplayskip{-1mm}
\begin{align}
k = \textbf{argmin}_{k} \sum_{I_*,l_*}\omega_* \left \| I_* - k\otimes l_*\right \|^2+\delta\left \| k \right \|^2\nonumber\\
(I_*,l_*) \in \{(\partial_x I, \partial_x l), (\partial_y I, \partial_y l),(\partial_{xx} I, \partial_{xx} l), \nonumber\\
(\partial_{yy} I, \partial_{yy} l),(\partial_{xy}I, (\partial_x\partial_y +\partial_y \partial_x)l/2)\}
\label{moBlur:eq:FMeng}
\end{align}
}

where $\delta$ denotes the weight of Tikhonov regularization and $\omega_* \in \{\omega_1,\omega_2\}$ represents a linear weight for the derivatives in different directions. Both $I$ and $l$ are propagated from the nearest coarse level within the pyramid. To minimise this energy~Eq.~(\ref{moBlur:eq:FMeng}), we follow the inner-iterative numerical scheme of~\cite{FMD} which yields a pre-estimated blur kernel $k$.

\subsection{Directional High-pass Filtering}
\label{moBlur:sec:filtering}

Once the pre-estimated kernel $k$ is obtained, our \emph{Directional High-pass Filters} are applied to enhance the blur information by reducing noise in the orthogonal direction of the tracked camera motion. Although our \emph{RGB-Motion Imaging System} provides an intuitively accurate camera motion estimation, outliers may still exist in the synchronisation. We take into account the directional components $\{\theta_1,\theta_2,\theta_{12}\}$ of two consecutive camera motions $M_1$ and $M_2$ as well as their combination $M_{12}$ (Fig.~\ref{moBlur:fig:camTrajectory}(d)) for extra robustness. The pre-estimated blur kernel is filtered along its orthogonal direction as follows:

\begin{align}
k = \sum_{\beta_*,\theta_*}\beta_*k\otimes f_{\theta_*+\pi/2}
\label{moBlur:eq:iterDirectional}
\end{align}

where $\beta_* \in \{1/2,1/3,1/6\}$ linearly weights the contribution of filtering in different directions. Note that two consecutive images $I_1$ and $I_2$ are involved in our framework where the former accepts the weight set $(\beta_*,\theta_*)\in \{(1/2,\theta_1), (1/3,\theta_2), (1/6,\theta_{12})\}$ while the other weight set $(\beta_*,\theta_*)\in \{(1/3,\theta_1), (1/2,\theta_2), (1/6,\theta_{12})\}$ is performed for the latter. This filtering process yields an updated blur kernel $k$ which is used to update the latent image $l$ within a non-blind deconvolution~\cite{Zhong}. Note that the convolution operation is computationally expensive in the spatial domain, we consider an equivalent filtering scheme in the frequency domain in the following subsection.

\subsection{Convolution for Directional Filtering}
\label{moBlur:sec:fastfiltering}

Our proposed directional filtering is performed as convolution operation in the spatial domain, which is often highly expensive in computation given large image resolutions. In our implementations, we consider a directional filtering scheme in the frequency domain where we have the equivalent form of filtering model Eq.~(\ref{moBlur:eq:filtering}) as follows:

\begin{equation}
K_\Theta(u,v) = K(u,v)F_\Theta(u,v)
\end{equation}

where $K_\Theta$ is the optimal blur kernel in the frequency domain while $K$ and $F_\Theta$ present the \emph{Fourier Transform} of the blur kernel $k$ and our directional filter $f_\theta$ respectively. Thus, the optimal blur kernel $k_\theta$ in the spatial domain can be calculated as $k_\theta=\texttt{IDFT}[K_\Theta]$ using \emph{Inverse Fourier Transform}. In this case, the equivalent form of our directional high-pass filter in the frequency domain is defined as follows:

\begin{equation}
F_\Theta(u,v) = 1-exp\left \{ -L^2(u,v)/2\sigma^{2} \right \}
\end{equation}

where the line function $L(u,v) = u \cos\theta+v \sin\theta$ controls the filtering process along the direction $\theta$ while $\sigma$ is the standard deviation for controlling the strength of the filter. Please note that other more sophisticated high-pass filters could also be employed using this directional substitution $L$. Even though this consumes a reasonable proportion of computer memory, convolution in the frequency domain $O(N\log_2N)$ is faster than equivalent computation in the spatial domain $O(N^2)$.

Having performed blind deconvolution and directional filtering (Sec.~\ref{moBlur:sec:blindDec}, \ref{moBlur:sec:filtering} and \ref{moBlur:sec:fastfiltering}), two updated blur kernels $k^i_1$ and $k^i_2$ on the $i$th level of the pyramid are obtained from input images $I^i_1$ and $I^i_2$ respectively, which is followed by the uniform blur image $b^i_1$ and $b^i_2$ computation using Eq.~(\ref{moBlur:eq:uniformImg}). In the following subsection, \emph{Blur-Robust Optical Flow Energy} optimisation on $b^i_1$ and $b^i_1$ is introduced in detail.

\subsection{Optical Flow Energy optimisation}
\label{moBlur:sec:optimisation}

As mentioned in Sec.~\ref{moBlur:sec:robustEng}, our blur-robust energy is continuous but highly nonlinear.
minimisation of such energy function is extensively studied in the optical flow community. In this section, a numerical scheme combining \emph{Euler-Lagrange Equations} and \emph{Nested Fixed Point Iterations} is applied~\cite{Brox} to solve our main energy function Eq.~\ref{moBlur:eq:Eng}. For clarity of presentation, we define the following mathematical abbreviations:

{\setlength\abovedisplayskip{-1mm}
\setlength\belowdisplayskip{-1mm}
\begin{eqnarray}
\begin{array}{ll}
b_{x}=\partial_{x}b_{2}(\textbf{x}+\textbf{w})& b_{yy}=\partial_{yy}b_{2}(\textbf{x}+\textbf{w}) \\
b_{y}=\partial_{y}b_{2}(\textbf{x}+\textbf{w}) &
b_{z}=b_{2}(\textbf{x}+\textbf{w})-b_{1}(\textbf{x})\\
b_{xx}=\partial_{xx}b_{2}(\textbf{x}+\textbf{w}) & b_{xz}=\partial_{x}b_{2}(\textbf{x}+\textbf{w})-\partial_{x}b_{1}(\textbf{x})\\
b_{xy}=\partial_{xy}b_2(\textbf{x}+\textbf{w}) & b_{yz}=\partial_{y}b_{2}(\textbf{x}+\textbf{w})-\partial_{y}b_{1}(\textbf{x})
\end{array}\nonumber
\end{eqnarray}
}

At the first phase of energy minimization, a system is built based on Eq.~\ref{moBlur:eq:Eng} where Euler-Lagrange is employed as follows:

{\setlength\abovedisplayskip{-1mm}
\setlength\belowdisplayskip{-1mm}
\begin{align}
\phi'\{b^2_z+\alpha(b^2_{xz}+b^2_{yz})\}\cdot \{b_xb_z+\alpha(b_{xx}b_{xz}+b_{xy}b_{yz})\}&\nonumber\\
-\gamma\phi'(\left \| \nabla u \right \|^2+\left \| \nabla v \right \|^2)\cdot \nabla u&=0\\
\phi'\{b^2_z+\alpha(b^2_{xz}+b^2_{yz})\}\cdot \{b_yb_z+\alpha(b_{yy}b_{yz}+b_{xy}b_{xz})\}&\nonumber\\
-\gamma\phi'(\left \| \nabla u \right \|^2+\left \| \nabla v \right \|^2)\cdot \nabla v&=0
\end{align}
}

An $n$-level image pyramid is then constructed from the top coarsest level to the bottom finest level. The flow field is initialized as $\textbf{w}^0=(0,0)^T$ on the top level and the outer fixed point iterations are applied on $\textbf{w}$. We assume that the solution $\textbf{w}^{i+1}$ converges on the $i+1$ level. We have:

{\setlength\abovedisplayskip{-1mm}
\setlength\belowdisplayskip{-1mm}
\begin{align}
\phi'\{(b^{i+1}_z)^2+\alpha(b^{i+1}_{xz})^2+\alpha(b^{i+1}_{yz})^2\}&\nonumber\\
\cdot\{b^i_xb^{i+1}_z+\alpha(b^i_{xx}b^{i+1}_{xz}+b^i_{xy}b^{i+1}_{yz})\}\nonumber\\
-\gamma\phi'(\left \| \nabla u^{i+1} \right \|^2+\left \| \nabla v^{i+1} \right \|^2)\cdot \nabla u^{i+1}&=0
\label{eq:FS_1}\\
\nonumber\\
\phi'\{(b^{i+1}_z)^2+\alpha(b^{i+1}_{xz})^2+\alpha(b^{i+1}_{yz})^2\}&\nonumber\\
\cdot\{b^i_yb^{i+1}_z+\alpha(b^i_{yy}b^{i+1}_{yz}+b^i_{xy}b^{i+1}_{xz})\}\nonumber\\
-\gamma\phi'(\left \| \nabla u^{i+1} \right \|^2+\left \| \nabla v^{i+1} \right \|^2)\cdot \nabla v^{i+1}&=0
\label{eq:FS_2}
\end{align}
}

Because of the nonlinearity in terms of $\phi'$, $b^{i+1}_*$, the system (Eqs.~\ref{eq:FS_1}, \ref{eq:FS_2}) is difficult to solve by linear numerical methods. We apply the first order Taylor expansions to remove these nonlinearity in $b^{i+1}_*$, which results in:

{\setlength\abovedisplayskip{-1mm}
\setlength\belowdisplayskip{-1mm}
\begin{align*}
b^{i+1}_{z} &\approx b^i_z+b^i_xdu^i+b^i_ydv^i, \\
b^{i+1}_{xz} &\approx b^k_{xz}+b^i_{xx}du^i+b^i_{xy}dv^i,\\
b^{i+1}_{yz} &\approx b^k_{yz}+b^i_{xy}du^i+b^i_{yy}dv^i.\\
\end{align*}
}

Based on the coarse-to-fine flow assumption of Brox~\emph{et al.}~\cite{Brox} w.r.t. $u^{i+1} \approx u^i + du^i$ and $v^{i+1} \approx v^i + dv^i$ where the unknown flow field on the next level $i+1$ can be obtained using the flow field and its incremental from the current level $i$. The new system can be presented as follows:

{\setlength\abovedisplayskip{-1mm}
\setlength\belowdisplayskip{-1mm}
 \begin{align}
(\phi')_B^{i}\cdot\{b_{x}^i(b_{z}^i+b_{x}^idu^{i}+b_{y}^idv^{i}) \nonumber \\
+\alpha b_{xx}^i(b_{xz}^i+b_{xx}^idu^{i}+b_{xy}^idv^{i})\nonumber\\
+\alpha b_{xy}^i(b_{yz}^i+b_{xy}^idu^{i}+b_{yy}^idv^{i})\} \nonumber \\
-\gamma(\phi')_S^{i}\cdot\nabla(u^i+du^{i})&=0
\label{eq:fixedK_1}\\
\nonumber\\
(\phi')_B^{i}\cdot\{b_{y}^i(b_{z}^i+b_{x}^idu^{i}+b_{y}^idv^{i}) \nonumber \\
+\alpha b_{yy}^i(b_{yz}^i+b_{xy}^idu^{i}+b_{yy}^idv^{i})\nonumber \\
+\alpha b_{xy}^i(b_{xz}^i+b_{xx}^idu^{i}+b_{xy}^idv^{i})\}\nonumber \\
-\gamma(\phi')_S^{i}\cdot\nabla(v^i+dv^{i})&=0
\label{eq:fixedK_2}
\end{align}
}

where the terms $(\phi')_{B}^{i}$ and $(\phi')_{S}^{i}$ contained $\phi$ provide robustness to flow discontinuity on the object boundary. In addition, $(\phi')_{S}^{i}$ is also regularizer for a gradient constraint in motion space. Although we fixed $\textbf{w}^i$ in Eqs.~\ref{eq:fixedK_1} and \ref{eq:fixedK_2}, the nonlinearity in $\phi'$ leads to the difficulty of solving the system. The inner fixed point iterations are applied to remove this nonlinearity: $du^{i,j}$ and $dv^{i,j}$ are assumed to converge within $j$ iterations by initializing $du^{i,0}=0$ and $dv^{i,0}=0$. Finally, we have the linear system in $du^{i,j+1}$ and $dv^{i,j+1}$ as follows:

{\setlength\abovedisplayskip{-1mm}
\setlength\belowdisplayskip{-1mm}
\begin{align}
(\phi')_B^{i,j}\cdot\{b_{x}^i(b_{z}^i+b_{x}^idu^{i,j+1}+b_{y}^idv^{i,j+1}) \nonumber \\ +\alpha b_{xx}^i(b_{xz}^i+b_{xx}^idu^{i,j+1}+b_{xy}^idv^{i,j+1})\nonumber\\
+\alpha b_{xy}^i(b_{yz}^i+b_{xy}^idu^{i,j+1}+b_{yy}^idv^{i,j+1})\} \nonumber \\
-\gamma(\phi')_S^{i,j}\cdot\nabla(u^i+du^{i,j+1})&=0\label{eq:EnergyKL_1}\\
\nonumber\\
(\phi')_B^{i,j}\cdot\{b_{y}^i(b_{z}^i+b_{x}^idu^{i,j+1}+b_{y}^idv^{i,j+1}) \nonumber \\ +\alpha b_{yy}^i(b_{yz}^i+b_{xy}^idu^{i,j+1}+b_{yy}^idv^{i,j+1})\nonumber\\
+\alpha b_{xy}^i(b_{xz}^i+b_{xx}^idu^{i,j+1}+b_{xy}^idv^{i,j+1})\}\nonumber \\
-\gamma(\phi')_S^{i,j}\cdot\nabla(v^i+dv^{i,j+1})&=0
\label{eq:EnergyKL_2}
\end{align}
}

where $(\phi')_{B}^{i,j}$ denotes a robustness factor against flow discontinuity and occlusion on the object boundaries. $(\phi')_{S}^{i,j}$ represents the diffusivity of the smoothness regularization.

{\setlength\abovedisplayskip{-1mm}
\setlength\belowdisplayskip{-1mm}
\begin{align}
(\phi')_{B}^{i,j}&=\phi'\{(b_{z}^i+b_{x}^idu^{i,j}+b_{y}^{i,j}dv^{i,j})^2 \nonumber \\
&+\alpha (b_{xz}^i+b_{xx}^idu^{i,j}+b_{xy}^{i}dv^{i,j})^2 \nonumber \\
&+\alpha (b_{yz}^i+b_{xy}^idu^{i,j}+b_{yy}^idv^{i,j})^2\} \nonumber \\
(\phi')_{S}^{i,j}&=\phi'\{\left \|\nabla(u^i+du^{i,j})\right \|^2+\left \|\nabla(v^i+dv^{i,j})\right \|^2\} \nonumber
\end{align}
}

In our implementation, the image pyramid is constructed with a downsampling factor of 0.75. The final linear system in Eq.~(\ref{eq:EnergyKL_1},\ref{eq:EnergyKL_2}) is solved using \emph{Conjugate Gradients} within 45 iterations.

\begin{table}[t!]
\centerline{
\begin{tabular}{l}
\hline
\hspace{5 pt}\textbf{Algorithm 2}: Auto Blur-Robust Optical Flow Framework\\
\hline
\hspace{5 pt}\textbf{Input} \hspace{6 pt} :~A image pair $I_1$, $I_2$ \emph{\textbf{Without}} camera motion\\
\hspace{5 pt}\textbf{Output}~:~Optimal optical flow field $\textbf{w}$\vspace{1pt}\\
\hspace{5 pt}1:\hspace{5 pt} \emph{A $n$-level top-down pyramid is built with the level index $i$}\\
\hspace{5 pt}2:\hspace{5 pt} $i \leftarrow 0$\\
\hspace{5 pt}3:\hspace{5 pt} $l^i_1 \leftarrow I^i_1$, $l^i_2 \leftarrow I^i_2$\\
\hspace{5 pt}4:\hspace{5 pt} $k^i_1 \leftarrow 0$, $k^i_2 \leftarrow 0$, $\textbf{w}^i\leftarrow (0,0)^T$, $\theta^i = 0$ \vspace{1pt}\\
\hspace{5 pt}5:\hspace{5 pt} \textbf{for} \emph{coarse to fine} \textbf{do}\\
\hspace{5 pt}6:\hspace{30 pt} $i \leftarrow i+1$\\
\hspace{5 pt}7:\hspace{30 pt} Resize $k^i_{\{1,2\}}$, $l^i_{\{1,2\}}$, $I^i_{\{1,2\}}$ and $\textbf{w}^i$ with the $i$th scale\\
\hspace{5 pt}8:\hspace{30 pt} \textbf{foreach} $* \in \{1,2\}$ \textbf{do}\\
\hspace{5 pt}9:\hspace{55 pt} $k^i_* \leftarrow$ \texttt{IterBlindDeconv} ( $l^i_*, I^i_*$ )\\
10:\hspace{55 pt} $k^i_* \leftarrow$ \texttt{DirectFilter} ( $k^i_*, \theta^i$ )\\
11:\hspace{55 pt} $l^i_* \leftarrow$ \texttt{NonBlindDeconvolve} ( $k^i_*, I^i_*$ )\\
12:\hspace{30 pt} \textbf{endfor}\\
13:\hspace{30 pt} $b^i_1 \leftarrow I^i_1 \otimes k^i_2$, $b^i_2 \leftarrow I^i_2 \otimes k^i_1$\\
14:\hspace{30 pt} $d\textbf{w}^i\leftarrow$ \texttt{EnergyOptimisation} ( $b^i_1, b^i_2, \textbf{w}^i$ )\\
15:\hspace{30 pt} $\textbf{w}^i\leftarrow \textbf{w}^i+d\textbf{w}^i$\\
16:\hspace{30 pt} $\theta^i\leftarrow$ \texttt{CameraMotionEstimation}($\textbf{w}^i$) \\
17:\hspace{5 pt} \textbf{endfor}\\
\hline
\label{moBlur:tab:autoKernelOflow}
\end{tabular}
\vspace{-3mm}
}
\end{table}

\subsection{Alternative Implementation with Automatic Camera Motion $\theta_*$ Estimation}
\label{moblur:sec:auto}

Alternative to using our assisted tracker, we also provide an additional implementation by using the camera motion $\theta_*$ estimated generically from the flow field. As shown in Algorithm 2, the system does not take the camera motion ($\theta_*$) as input but computes it (\texttt{CameraMotionEstimation}) generically at every level of the image pyramid.

{\setlength\abovedisplayskip{-1mm}
\setlength\belowdisplayskip{-1mm}
\begin{align}
\textbf{A}^i&\leftarrow \texttt{AffineEstimation}(\textbf{x},\textbf{x}+\textbf{w}^i)\nonumber\\
\theta^i&\leftarrow \texttt{AffineToMotionAngle}(\textbf{A}^i)
\end{align}
}

On each level, we calculate the \emph{Affine} Matrix from $I_1^i$ to $I_2^i$ using the correspondences $\textbf{x} \rightarrow \textbf{x}+\textbf{w}^i$ and RANSAC. The translation information from $\textbf{A}^i$ is then normalized and converted to the angle format $\theta^i$. In this case, our $DirectionalFilter$ is also downgraded to consider one direction $\theta^i$ for each level. In the next section, we quantitatively compare our method to other popular baselines.

\section{Evaluation}
\label{moBlur:sec:evaluation}

\begin{figure*}[t!]
\centerline{
\includegraphics[width=0.9\linewidth]{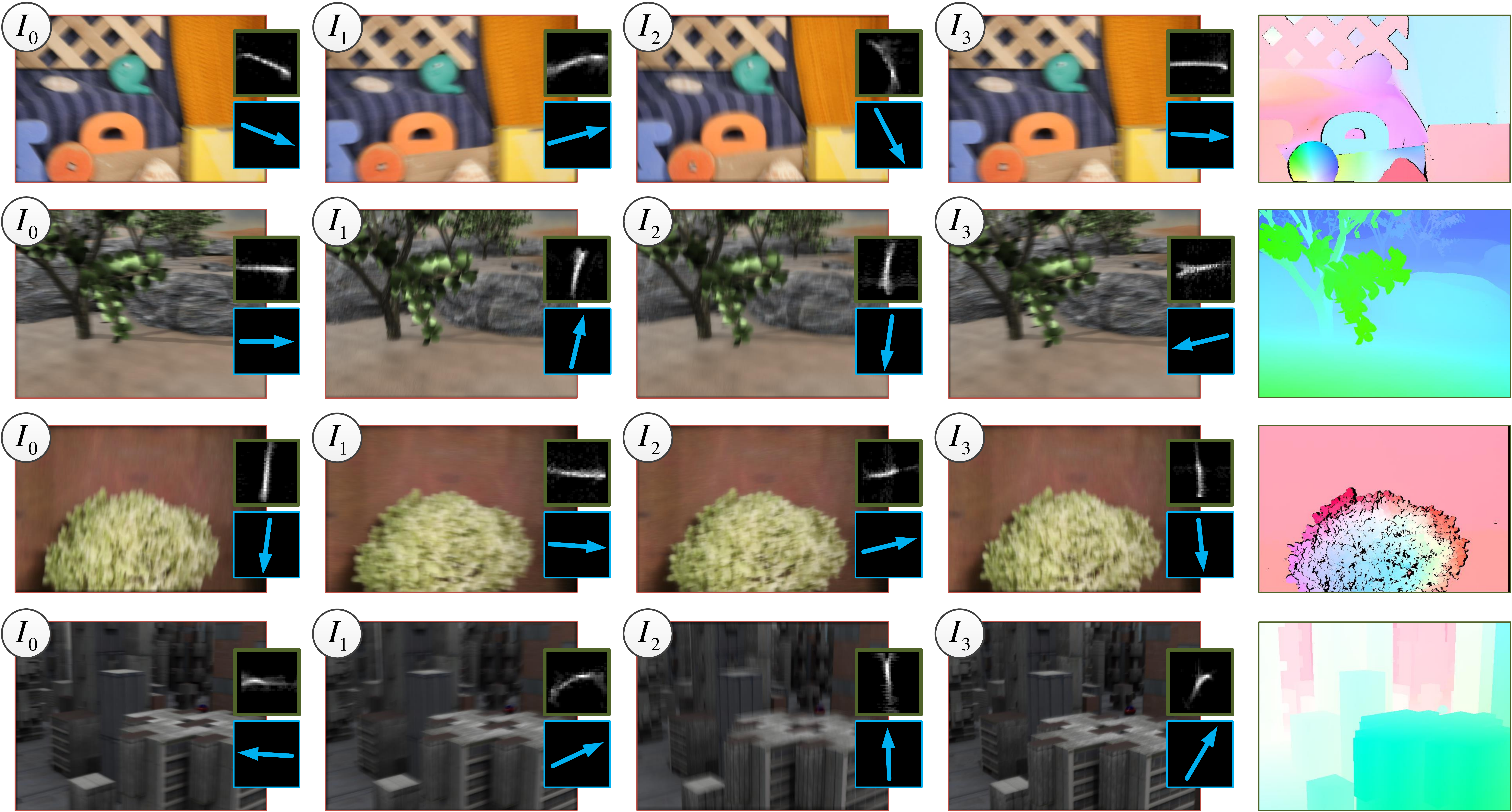}\vspace{-2mm}
}
\caption{The synthetic blur sequences with the blur kernel, tracked camera motion direction and ground truth flow fields. \textbf{From Top To Bottom}: sequences of \emph{RubberWhale}, \emph{Urban2}, \emph{Hydrangea} and \emph{Urban2}.}
\label{moBlur:fig:evaSynSample}
\end{figure*}

In this section, we evaluate our method on both synthetic and realworld sequences and compare its performance against three existing state-of-the-art optical flow approaches of Xu \emph{et al.}'s MDP~\cite{Xu_deblur}, Portz \emph{et al.}'s~\cite{Portz} and Brox~\emph{et al.}'s~\cite{Brox} (an implementation of~\cite{liu}). MDP is one of the best performing optical flow methods given blur-free scenes, and is one of the top 3 approaches in the Middlebury benchmark~\cite{Middlebury}. Portz \emph{et al.}'s method represents the current state-of-the-art in optical flow estimation given object blur scenes while Brox~\emph{et al.}'s contains a similar optimisation framework and numerical scheme to Portz \emph{et al.}'s, and ranks in the midfield of the Middlebury benchmarks based on overall average. Note that all three baseline methods are evaluated using their default parameters setting; all experiments are performed using a 2.9Ghz Xeon 8-cores, NVIDIA Quadro FX 580, 16Gb memory computer.

In the following subsections, we compare our algorithm (\emph{moBlur}) and four different implementations (\emph{auto}, \emph{nonGC}, \emph{nonDF} and \emph{nonGCDF}) against the baseline methods. \emph{auto} denotes the implementation using the automatic camera motion estimation scheme (Algorithm 2); \emph{nonGC} represents the implementation \textbf{without} the \emph{Gradient Constancy} term while \emph{nonDF} denotes an implementation \textbf{without} the directional filtering process. \emph{nonGCDF} is the implementation with neither of these features. The results show that our \emph{Blur-Robust Optical Flow Energy} and \emph{Directional High-pass Filter} significantly improve algorithm performance for blur scenes in both synthetic and realworld cases.


\subsection{Middlebury Dataset with camera shake blur}


One advance for evaluating optical flow given scenes with object blur is proposed by Portz \emph{et al.}~\cite{Portz} where synthetic \emph{Ground Truth} (GT) scenes are rendered with blurry moving objects against a blur-free static/fixed background. However, their use of synthetic images and controlled object trajectories lead to a lack of global camera shake blur, natural photographic properties and real camera motion behaviour. To overcome these limitations, we render four sequences with camera shake blur and corresponding GT flow-fields by combining sequences from the Middlebury dataset~\cite{Middlebury} with blur kernels estimated using our system.

\begin{figure*}[t!]
    \centerline{
    \subfigure[\textbf{Left}: Quantitative \emph{Average Endpoint Error} (AEE), \emph{Average Angle Error} (AAE) and \emph{Time Cost} (in second) comparisons on our synthetic sequences where the subscripts show the rank in relative terms. \textbf{Right}: AEE measure on \emph{RubberWhale} by ramping up the noise distribution. ]{\includegraphics[width=0.95\linewidth]{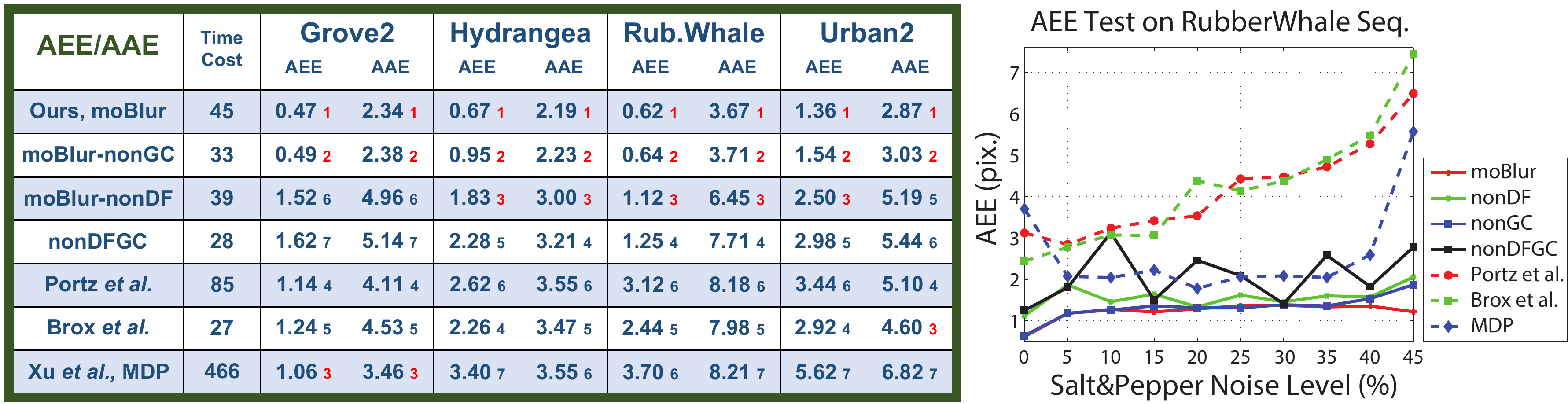}\label{moBlur:fig:evaSynTest}}
    }
    \centerline{
    \subfigure[Visual comparison on sequences \emph{RubberWhale}, \emph{Urban2}, \emph{Hydrangea} and \emph{Urban2} by varying baseline methods. For each sequence, \textbf{First Row}: optical flow fields from different methods. \textbf{Second Row}: the error maps against the ground truth. ]{\includegraphics[width=0.95\linewidth]{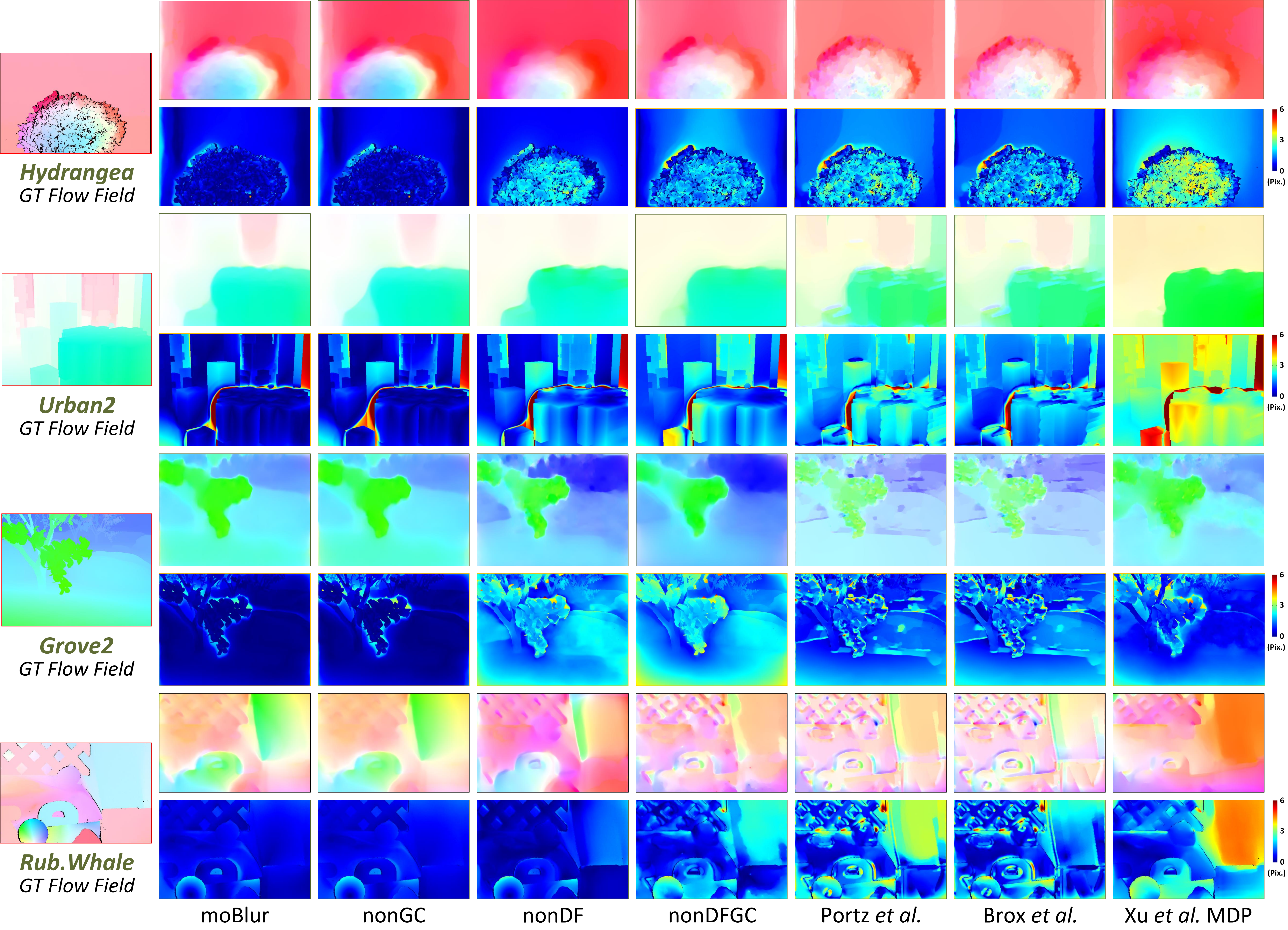}\label{moBlur:fig:evaSynFlow}} 
    }
    \caption{Quantitative evaluation on four synthetic blur sequences with both camera motion and ground truth.}
    \label{moBlur:fig:evaSyn}
\end{figure*}

In our experiments we select the sequences \emph{Grove2}, \emph{Hydrangea}, \emph{RubberWhale} and \emph{Urban2} from the Middlebury dataset. For each of them, four adjacent frames are selected as latent images along with the GT flow field $\textbf{w}_{gt}$ (supplied by Middlebury) for the middle pair. $40\times 40$ blur kernels are then estimated~\cite{FMD} from realworld video streams captured using our \emph{RGB-Motion Imaging System}. As shown in Fig.~\ref{moBlur:fig:evaSynSample}, those kernels are applied to generate blurry images denoted by $I_0$, $I_1$, $I_2$ and $I_3$ while the camera motion direction is set for each frame based on the 3D motion data. Although the $\textbf{w}_{gt}$ between latent images can be utilised for the evaluation on relative blur images $I_*$~\cite{Sintel,SintelWS}, strong blur can significantly violate the original image intensity, which leads to a multiple correspondences problem: a point in the current image corresponds to multiple points in the consecutive image. To remove such multiple correspondences, we sample reasonable correspondence set $\{\hat{\textbf{w}}~|~\hat{\textbf{w}}\subset\textbf{w}_{gt}, \left | I_2(\textbf{x}+\hat{\textbf{w}})-I_1(\textbf{x})\right | < \epsilon\}$ to use as the GT for the blur images $I_*$ where $\epsilon$ denotes a predefined threshold. Once we obtain $\hat{\textbf{w}}$, both \emph{Average Endpoint Error} (AEE) and \emph{Average Angle Error} (AAE) tests~\cite{Middlebury} are considered in our evaluation. The computation is formulated as follows:

{\setlength\abovedisplayskip{-1mm}
\setlength\belowdisplayskip{-1mm}
\begin{align}
AEE &= \frac{1}{n} \sum_{\textbf{x}}\sqrt{(u-\hat{u})^2+(v-\hat{v})^2}\\
AAE &= \frac{1}{n} \sum_{\textbf{x}} \cos^{-1} \left (  \frac{1.0+u \times \hat{u}+v \times \hat{v}}{\sqrt{1.0+u^2+v^2} \sqrt{1.0+\hat{u}^2+\hat{v}^2}} \right )
\end{align}
}

\begin{figure*}[t!]
\centerline{
\includegraphics[width=0.97\linewidth]{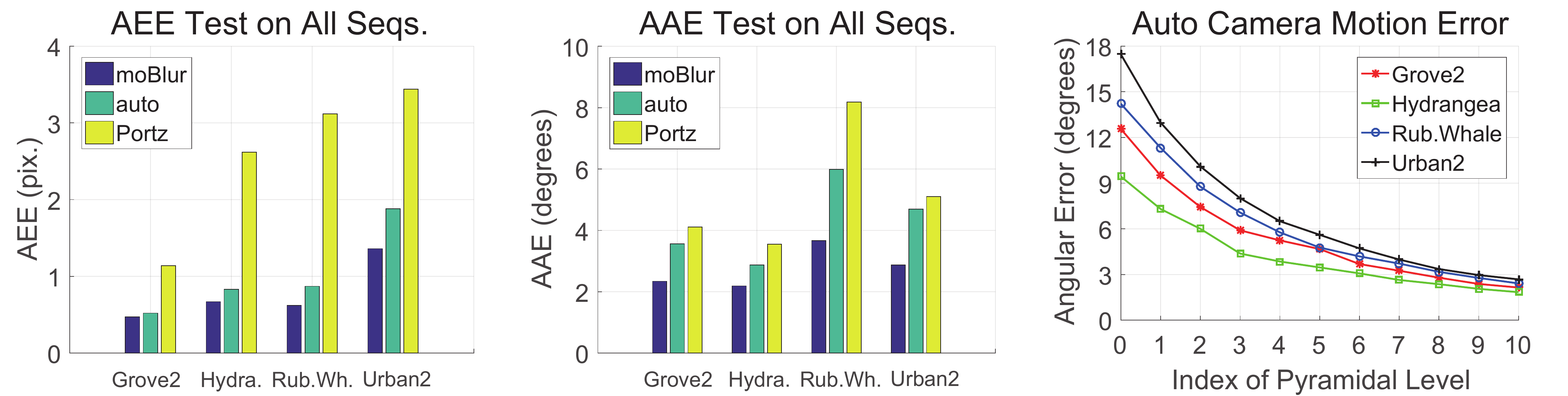}
}
\caption{Quantitative comparison between our implementations using \emph{RGB-Motion Imaging} (\emph{\textbf{moBlur}}); and automatic camera motion estimation scheme (\emph{\textbf{auto}}, see Sec.~\ref{moblur:sec:auto}). \textbf{From Left To Right}: AEE and AAE tests on all sequences respectively; the angular error of camera motion estimated by \textbf{\emph{auto}} by varying the pyramidal levels of the input images.}
\label{moBlur:fig:angAuto}
\end{figure*}

\begin{figure*}[t!]
\centerline{
\includegraphics[width=0.97\linewidth]{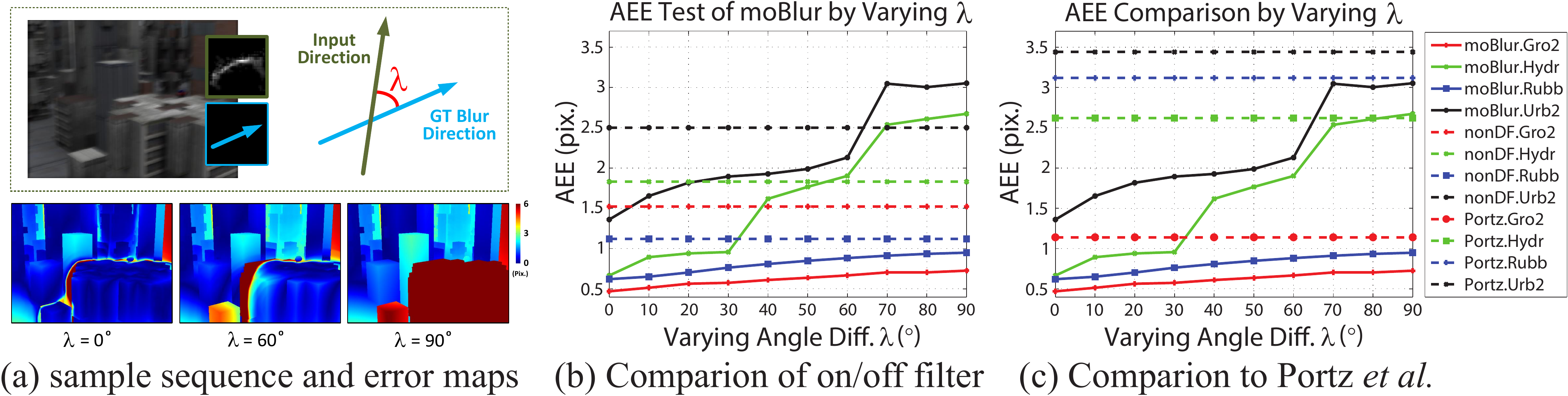}
}
\caption{AEE measure of our method (\emph{moBlur}) by varying the input motion directions. \textbf{(a)}: the overall measure strategy and error maps of \emph{moBlur} on sequence \emph{Urban2}. \textbf{(b)}: the quantitative comparison of \emph{moBlur} against \emph{nonDF} by ramping up the angle difference $\lambda$. \textbf{(c)}: the measure of \emph{moBlur} against Portz \emph{et al.}~\cite{Portz}.}
\label{moBlur:fig:rampDF}
\end{figure*}

where $\textbf{w}=(u,v)^T$ and $\hat{\textbf{w}}=(\hat{u},\hat{v})^T$ denotes the baseline flow field and the ground truth flow field (by removing multiple correspondences) respectively while $n$ presents the number of ground truth vectors in $\hat{\textbf{w}}$. The factor $1.0$ in AAE is an arbitrary scaling constant to convert the units from pixels to degrees~\cite{Middlebury}. Fig.~\ref{moBlur:fig:evaSynTest} \emph{Left} shows AEE (in pixel) and AAE (in degree) tests on our four synthetic sequences. \emph{moBlur} and \emph{nonGC} lead both AEE and AAE tests in all the trials. Both Brox \emph{et al.} and MDP yield significant error in \emph{Hydrangea}, \emph{RubberWhale} and \emph{Urban2} because those sequences contain large textureless regions with blur, which in turn weakens the inner motion estimation process as shown in Fig.~\ref{moBlur:fig:evaSynFlow}. Fig.~\ref{moBlur:fig:evaSynTest} also illustrates the average time cost (second per frame) of the baseline methods. Our method gives reasonable performance (45 sec. per frame) comparing to the state-of-the-art Portz \emph{et al.} and MDP even an inner image deblurring process is involved. Furthermore, Fig~\ref{moBlur:fig:evaSynTest} \emph{Right} shows the AEE metric for \emph{RubberWhale} by varying the distribution of \emph{Salt\&Pepper} noise. It is observed that a higher noise level leads to additional errors for all the baseline methods. Both \emph{moBlur} and \emph{nonGC} yield the best performance while Portz \emph{et al.} and Brox \emph{et al.} show a similar rising AEE trend when the noise increases.

\begin{figure*}[t!]
\centerline{
\includegraphics[width=0.9\linewidth]{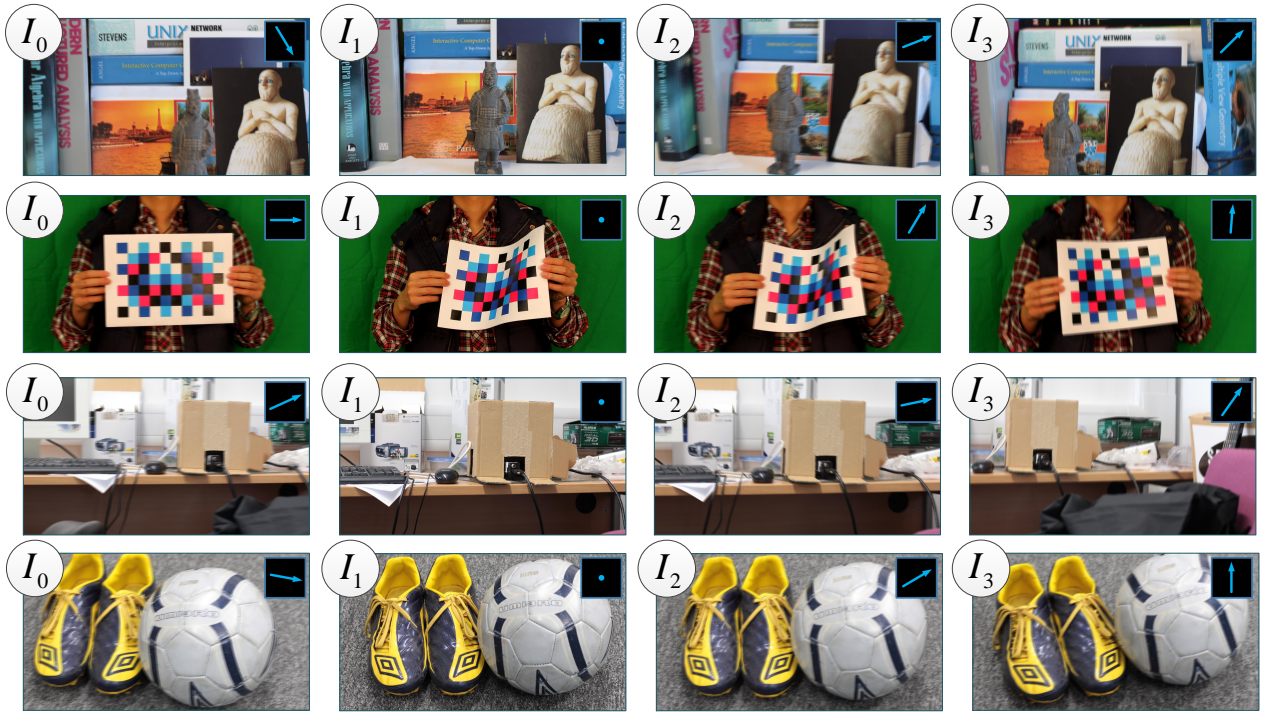}
}
\caption{The realworld sequences captured along the tracked camera motion. \textbf{From Top To Bottom}: sequences of \emph{warrior}, \emph{chessboard}, \emph{LabDesk} and \emph{shoes}.}
\label{moBlur:fig:realSamples}
\end{figure*}

Fig.~\ref{moBlur:fig:angAuto} shows our quantitative measure by comparing our two implementations which use the \emph{RGB-Motion Imaging} (\textbf{\emph{moBlur}}) and automatic camera motion estimation scheme (\textbf{\emph{auto}}, see Sec.~\ref{moblur:sec:auto}) respectively. For better observation, we also give the Portz~\emph{et al.} in this measure. We observe that both our implementations outperform Portz~\emph{et al.} in the AEE and AAE tests. Especially the \textbf{\emph{moBlur}} gives the best accuracy in all trials. The implementation \textbf{\emph{auto}} yields the less accurate results than the \textbf{\emph{moBlur}}. It may be because the auto camera motion estimation is affected by ambiguous blur that often caused by multiple moving objects. To investigate this issue, we plot the angular error by comparing the auto-estimated camera motion to the ground truth on all the sequences (Fig.~\ref{moBlur:fig:angAuto}, right end). We observe that our automatic camera motion estimation scheme leads to higher errors on the upper/coarser level of the image pyramid. Even the accuracy is improved on the finer levels but the error may be accumulated and affect the final result.

In practice, the system may be used in some challenge scenes, e.g. fast camera shaking, super high frame rate capture, or even infrared interference, etc. In those cases, the wrong tracked camera motion may be given to some specific frames. To investigate how the tracked camera motion affects the accuracy of our algorithm, we compare \emph{moBlur} to \emph{nonDF} (our method without directional filtering) and Portz \emph{et al.} by varying the direction of input camera motion. As shown in Fig.~\ref{moBlur:fig:rampDF}(a), we rotate the input camera motion vector with respect to the GT blur direction by an angle of $\lambda$ degrees. Here $\lambda=0$ represents the ideal situation where the input camera motion has the same direction as the blur direction. The increasing $\lambda$ simulates more errors in the camera motion estimation. Fig.~\ref{moBlur:fig:rampDF}(b,c) shows the AEE metric by increasing the $\lambda$. We observe that the AEE increases during this test. \emph{moBlur} outperforms the \emph{nonDF} (\emph{moBlur} without the directional filter) in both \emph{Grove2} and \emph{RubberWhale} while \emph{nonDF} provides higher performance in \emph{Hydrangea} when $\lambda$ is larger than $50^\circ$. In addition, \emph{moBlur} outperforms Portz \emph{et al.} in all trials except \emph{Hydrangea} where Portz \emph{et al.} shows a minor advantage (AEE 0.05) when $\lambda=90^\circ$. The rationale behind this experiment is that the wrong camera motion may yield significant information loss in the directional high-pass filtering. Such information loss harms the deblurring process and consequently leads to errors in the optical flow estimation. Thus, obtaining precise camera motion is the essential part of this system, as well as a potential future research.

\subsection{Realworld Dataset}

\begin{figure*}[t!]
    \centerline{
    \subfigure[Visual comparison on realworld sequences of \emph{warrior} and \emph{chessboard}. ]{\includegraphics[width=0.95\linewidth]{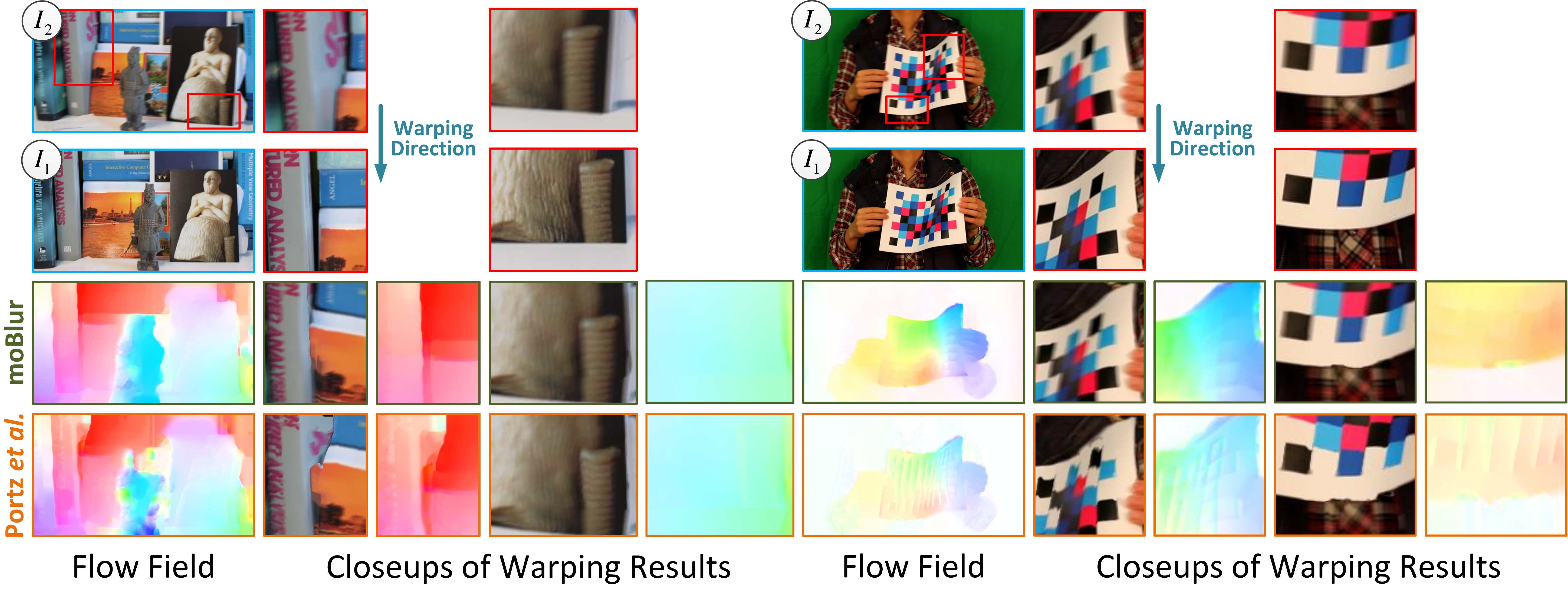}}
    }

    \centerline{
    \subfigure[Visual comparison on realworld sequences of \emph{LabDesk} and \emph{shoes}.\vspace{-2mm} ]{\includegraphics[width=0.95\linewidth]{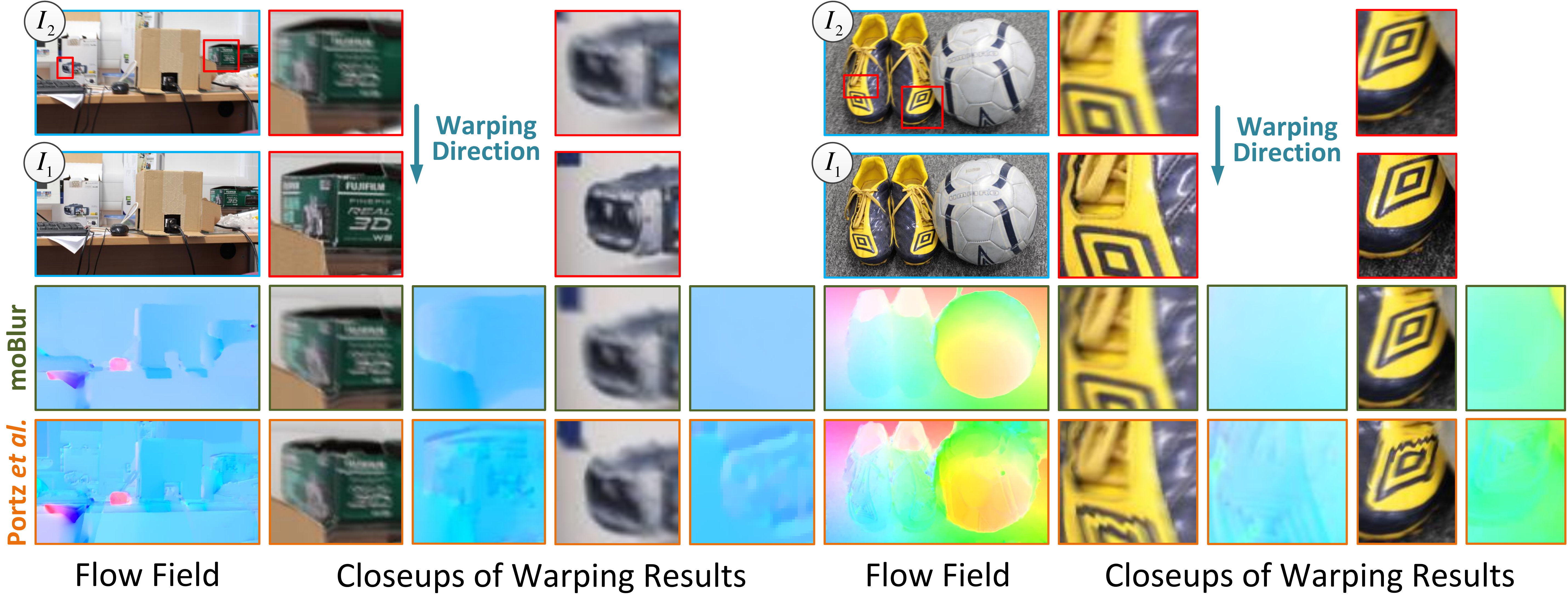}}
    }
    \caption{Visual comparison of image warping on realworld sequences of \emph{warrior}, \emph{chessboard}, \emph{LabDesk} and \emph{shoes}, captured by our \emph{RGB-Motion Imaging System}.}
    \label{moBlur:fig:evaReal}
\end{figure*}

To evaluate our method in the realworld scenes, we capture four sequences \emph{warrior}, \emph{chessboard}, \emph{LabDesk} and \emph{shoes} with tracked camera motion using our \emph{RGB-Motion Imaging System}. As shown in Fig.~\ref{moBlur:fig:realSamples}, both \emph{warrior} and \emph{chessboard} contain occlusions, large displacements and depth change while the sequences of \emph{LabDesk} and \emph{shoes} embodies the object motion blur and large textureless regions within the same scene. Fig.~\ref{moBlur:fig:evaReal} shows visual comparison of our method \emph{moBlur} against Portz \emph{et al.} on these realworld sequences. It is observed that our method preserves appearance details on the object surface and reduce boundary distortion after warping using the flow field. In addition, our method shows robustness given cases where multiple types of blur exist in the same scene (Fig.\ref{moBlur:fig:evaReal}(b), sequence \emph{shoes}).

\section{Conclusion}

In this paper, we introduce a novel dense tracking framework which interleaves both a popular iterative blind deconvolution; as well as a warping based optical flow scheme. We also investigate the blur papameterization for the video footages. In our evaluation, we highlight the advantages of using both the extra motion channel and the directional filtering in the optical flow estimation for the blurry video footages. Our experiments also demonstrated the improved accuracy of our method against large camera shake blur in both noisy synthetic and realworld cases. One limitation in our method is that the spatial invariance assumption for the blur is not valid in some realworld scenes, which may reduce accuracy in the case where the object depth significantly changes. Finding a depth-dependent deconvolution and deep data-driven model would be a challenge for future work as well.

\section{Acknowledgements}

We thank Ravi Garg and Lourdes Agapito for providing their GT datasets. We also thank Gabriel Brostow and the UCL Vision Group for their helpful comments. The authors are supported by the EPSRC CDE EP/L016540/1 and CAMERA EP/M023281/1; and EPSRC projects EP/K023578/1 and EP/K02339X/1.



\section*{References}
\bibliographystyle{elsarticle-num}
\bibliography{bib}





\end{document}